\documentclass[letterpaper, 10 pt, conference]{ieeeconf}  

\IEEEoverridecommandlockouts                              

\usepackage[usenames,dvipsnames,table,xcdraw]{xcolor}
\usepackage{cite}
\usepackage{graphicx}
\usepackage{makecell, multirow}
\usepackage{amsfonts}
\usepackage{amsmath}
\usepackage{hyperref}
\usepackage{siunitx}
\usepackage{tikz}

\graphicspath{ {./figures/} }

\DeclareMathOperator*{\argmin}{arg\,min}

\title{\LARGE \bf
Adversarial Motion Priors Make Good Substitutes for \\Complex Reward Functions
}

\author{
\authorblockN{Alejandro Escontrela$^{\gamma, \sigma}$, Xue Bin Peng$^{\gamma}$, Wenhao Yu$^{\sigma}$, Tingnan Zhang$^{\sigma}$}
\vspace{0.03in}
\authorblockN{Atil Iscen$^{\sigma}$, Ken Goldberg$^{\gamma}$, Pieter Abbeel$^{\gamma}$}
\vspace{0.03in}
\authorblockA{{$\gamma$}: UC Berkeley, {$\sigma$}: Google Brain}
\vspace{0.05in}
\authorblockA{{$\gamma$}: \{escontrela, xbpeng, goldberg, pabbeel\}@berkeley.edu}
\authorblockA{{$\sigma$}: \{magicmelon, tingnanzhang, atil\}@google.com}
}

\begin{document}

\maketitle
\global\csname @topnum\endcsname 0
\global\csname @botnum\endcsname 0
\thispagestyle{empty}
\pagestyle{empty}


\begin{abstract}
Training a high-dimensional simulated agent with an under-specified reward function often leads the agent to learn physically infeasible strategies that are ineffective when deployed in the real world. To mitigate these unnatural behaviors, reinforcement learning practitioners often utilize complex reward functions that encourage physically plausible behaviors. However, a tedious labor-intensive tuning process is often required to create hand-designed rewards which might not easily generalize across platforms and tasks. We propose substituting complex reward functions with ``style rewards" learned from a dataset of motion capture demonstrations. A learned style reward can be combined with an arbitrary task reward to train policies that perform tasks using naturalistic strategies. These natural strategies can also facilitate transfer to the real world. We build upon Adversarial Motion Priors -- an approach from the computer graphics domain that encodes a style reward from a dataset of reference motions -- to demonstrate that an adversarial approach to training policies can produce behaviors that transfer to a real quadrupedal robot without requiring complex reward functions. We also demonstrate that an effective style reward can be learned from a few seconds of motion capture data gathered from a German Shepherd and leads to energy-efficient locomotion strategies with natural gait transitions.
\end{abstract}

\begin{figure}
    \centering
    \includegraphics[width=0.48\textwidth]{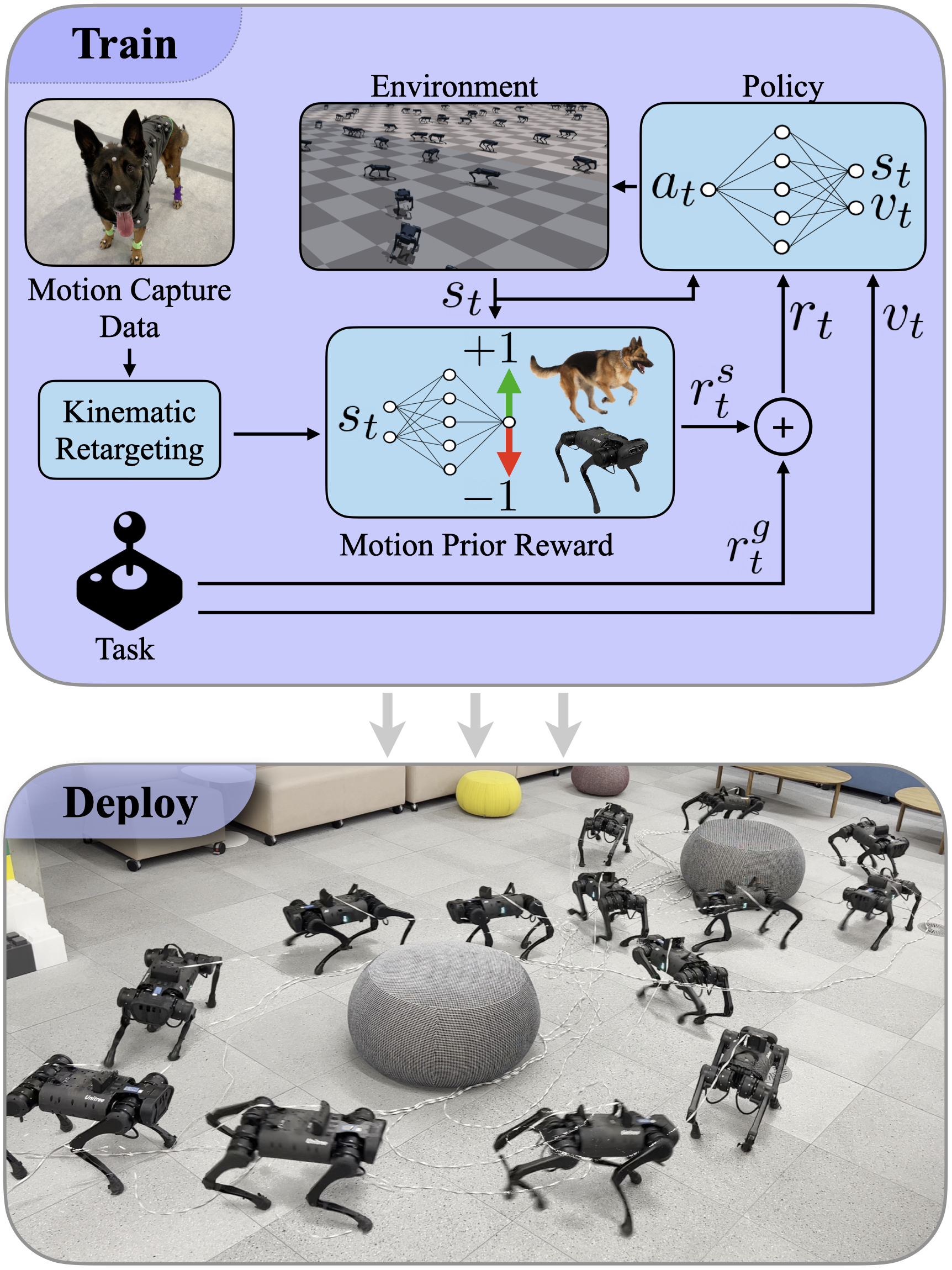}
    \caption{Training with Adversarial Motion Priors encourages the policy to produce behaviors which capture the essence of the motion capture dataset while satisfying the auxiliary task objective. Only a small amount of motion capture data is required to train the learning system (4.5 seconds in our experiments).}
    \label{fig:algorithm}
\end{figure}

\section{INTRODUCTION}

Developing controllers for high-dimensional continuous control systems such as legged robots has long been an area of study.
Early work in this field focused on developing approximate dynamics models of a system and then using trajectory optimization algorithms to solve for the actions that lead an agent to achieving a desired goal \cite{winkler18, raibert1986, xie20, Byl09}. However, the resulting controllers tend to be highly specialized for a particular task, limiting their ability to generalize across more diverse tasks or environments. More recently, there has been a surge in algorithms that use reinforcement learning (RL) to learn locomotion behaviors\cite{miki2022learning, lee20, kumar21, siekmann2021sim, Ji22}. This approach proved highly effective in simulation \cite{heess17}, but this success did not translate to the real world due to challenges associated with overcoming the simulation to reality gap.


\begin{figure*}[ht!]
    \includegraphics[width=\textwidth]{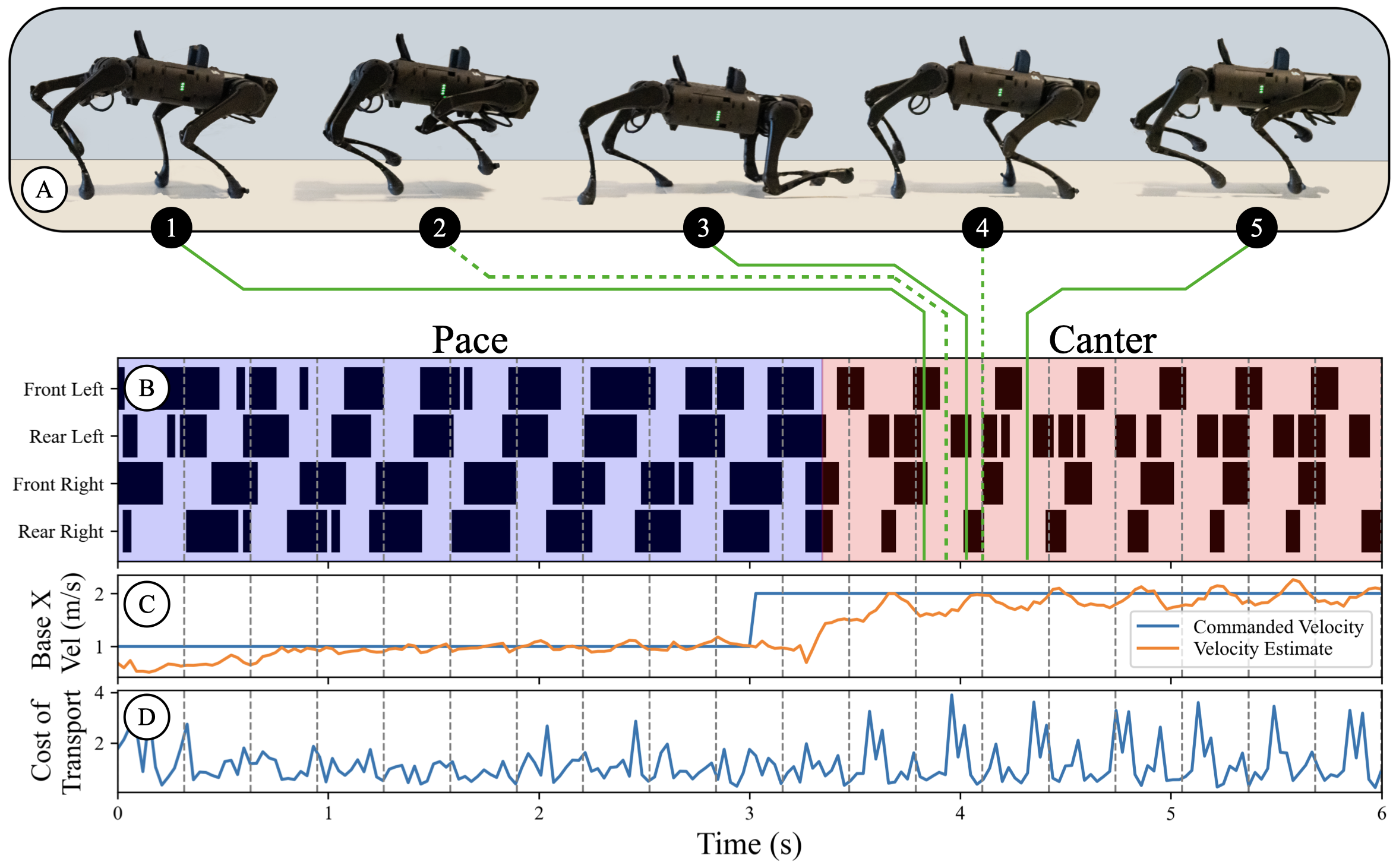}
    \caption{Key frames, gait pattern, velocity tracking, and energy-efficiency of the robot dog throughout a trajectory \textbf{A}: Key frames of A1 during a  canter motion overlaid on a plain background for contrast. \textbf{B}: Gait diagram indicating contact timing and duration for each foot in black. Training with Adversarial Motion Priors enables the policy to synthesize behaviors which lead to natural gait transitions at different velocities. \textbf{C}: Plot of commanded forward velocities and estimated velocities during the rollout. \textbf{D}: Estimated Cost of Transport (COT) during the rollout. While pacing the COT remains constant with small oscillations. However, when the robot enters a canter phase the COT exhibits spikes corresponding to the robot pushing off its hind legs and troughs corresponding to the flight phase where energy consumption is low. This gait transition phenomenon closely relates to the behavior of quadrupedal mammals, which modulate their gait according to their speed of travel, leading to minimal energy consumption consumption \cite{Hoyt1981}.}
    \label{fig:banner_canter}
\end{figure*}

One of the main challenges inhibiting RL approaches from being more effective in the real world is related to the aggressive and overly-energetic behaviors that are often learned by RL agents trained using under-specified reward functions. As an example, a legged RL agent trained with a reward that encourages forward velocity will often learn a control policy that exploits flailing of the limbs or high-impulse contacts, and other inaccurate simulator dynamics, to achieve forward movement. Such behaviors are unlikely to be effective when transferred to a real robot due to actuator limits and potential damage to the robot. To overcome the issues posed by reward under-specification, researchers have investigated task-specific action spaces \cite{iscen2019, yang21}, complex style reward formulations \cite{kumar21, lee20, miki22, siekmann2021sim}, and curriculum learning \cite{xie2020learning, yu2018learning}. These approaches achieve state-of-the-art results in locomotion, but defining custom action spaces and hand-designed reward functions requires substantial domain knowledge and a delicate tuning process. Additionally, these approaches are often platform-specific and do not generalize easily across tasks.

In the realm of computer graphics, \emph{Adversarial Motion Priors} (AMP) \cite{peng21} leverage GAN-style training to learn a ``style" reward from a reference motion dataset. The style reward encourages the agent to produce a trajectory distribution that minimizes the Pearson divergence between the reference trajectories and the policy trajectories\cite{mao17}. A simple task-specific reward can then be specified in conjunction with the style reward to produce policies that match the style of the dataset while performing the specified task (Fig. \ref{fig:algorithm}). Animators have leveraged this flexible approach to animate characters that perform complex and highly dynamic tasks while remaining human-like. 
However, the viability of this approach to train policies for the real world has not been studied, even though it could provide a promising alternative to the hand-defined complex rewards that are prevalent in recent literature \cite{lee20, kumar21}. In this work, we substitute complex hand-specified style reward formulations with a motion prior learned from a few seconds of German Shepherd motion capture data. We propose the following contributions:


\begin{itemize}
    \item We introduce a learning framework that leverages small amounts of motion capture data to encode a style reward that -- when trained in conjunction with an auxiliary task objective -- produces policies that can be effectively deployed on a real robot.
    \item 
    We study the energy efficiency  of agents trained with complex style reward formulations \cite{Rudin21, miki2022learning, lee20} and policies trained with Adversarial Motion Priors. We find that training policies with motion priors results in a lower Cost of Transport, and analyze the benefits of leveraging the energy-efficient prior provided by the data. We also find that policies trained with motion priors produce natural gait transitions which result in more energy-efficient motions across different speeds\footnote{Videos and the project repository can be found at: \url{https://bit.ly/3hpvbD6}}.
\end{itemize}


\section{RELATED WORK}

\subsection{Deep Reinforcement Learning for Robot Control}

Recent works in robotics have shown promising results in applying Deep Reinforcement Learning (DRL) to a variety of robotic control tasks such as manipulation\cite{akkaya2019solving, James16, Levine16, Gu16}, locomotion\cite{miki2022learning, siekmann2021sim, Ji22, lee20}, and navigation\cite{zhu2017target, fu2021}. DRL provides an effective paradigm for automatically synthesizing control policies for a given objective function, thus avoiding the need for manually designed controllers. However, controllers trained using DRL often lead to jerky, unnatural behaviors that may maximize the objective function, but may not be suitable for real-robot deployment \cite{heess17, ibarz2018reward}. As a result, manually-designed priors are often required to regularize the policy's behavior. For example, legged locomotion researchers have investigated complex reward functions \cite{miki2022learning, siekmann2021sim, lee2019robust}, task-specific action spaces \cite{iscen2019, tan2018sim, iscen21, yang21}, or curriculum learning \cite{xie2020learning, yu2018learning} to encourage robot behavior that is amenable to physical deployment. Despite the compelling results in these works, these approaches are often task-specific and require substantial effort to tune for each skill of interest. In this work, we explore the idea of automatically learning these behavioral priors directly from reference motion data.

\subsection{Motion Imitation}
Imitating reference motion data provides a general approach for developing controllers for a wide range of skills that would otherwise be difficult to manually encode into controllers \cite{Pollard2002,GrimesCR06,Suleiman08,Yamane2010}. These techniques often utilize some form of motion tracking, where a controller imitates a desired motion by explicitly tracking the sequence of target poses specified by a reference trajectory \cite{Atkeson1991,Nakaoka2003,KimHumanoid09,Koenemann2014RealtimeIO}. In simulated domains, motion tracking combined with reinforcement learning has been shown to be highly effective for reproducing a large corpus of complex and dynamic motor skills \cite{peng18,BasketballLiu2018,2018-TOG-SFV,MuscleConLee2019}. While motion tracking can be very effective for imitating individual motion clips, the tracking objective tends to constrain a controller to closely follow the reference motion, which can limit an agent's ability to develop more versatile and diverse behaviors as needed to fulfill auxiliary task objectives. Furthermore, it can be difficult to apply tracking-based techniques to imitate behaviors from diverse motion datasets, often requiring substantial overhead in the form of motion planners and task-specific annotation of the motion clips \cite{MocapImitationChentanez2018,DreCon2019,PredictSimPark2019,li21}. In this work, we utilize a more flexible motion imitation approach based on adversarial imitation learning, which allows our system to shape the behavior of an agent using unstructured motion datasets, while also affording the agent more flexibility to develop new behaviors as needed to achieve task objectives.

\subsection{Adversarial Imitation Learning} \label{section:AIL}
Adversarial imitation learning provides a flexible and scalable approach for imitating behaviors from diverse demonstration datasets (e.g. reference motions) \cite{ApprenticeAbbeel2004,MaxEntIRL2008,GAIL2016}. Rather than explicitly tracking individual motion clips, adversarial techniques aim to learn policies that match the state/trajectory distribution of the dataset \cite{FGAN2016,FDivKe2019}, which can provide the agent more flexibility in composing and interpolating between behaviors shown in the dataset. This is done by training an adversarial discriminator to differentiate between behaviors produced by a policy and behaviors depicted in the demonstration data. The discriminator then serves as the style reward for training a control policy to imitate the demonstrations. While these methods have shown promising results in low-dimensional domains \cite{GAIL2016,infogail2017}, when applied to high-dimensional continuous control tasks, the quality of the results produced by these methods generally falls well behind state-of-the-art tracking-based techniques \cite{Merel2017,DiverseImitationWang2017}. Recently, \emph{Peng et al.} \cite{peng21} proposed \emph{Adversarial Motion Priors} (AMP), which combines adversarial imitation learning with auxiliary tasks objectives, thereby enabling simulated agents to perform high-level tasks, while imitating behaviors from large unstructured motion datasets. We will leverage this adversarial technique to learn locomotion skills for legged robots. We show that the learned motion prior leads to more natural, physically plausible, and energy-efficient behaviors, which are then more amenable to transfer from simulation to a real-world robot.


\begin{figure*}
    \centering
    
    \includegraphics[width=\textwidth]{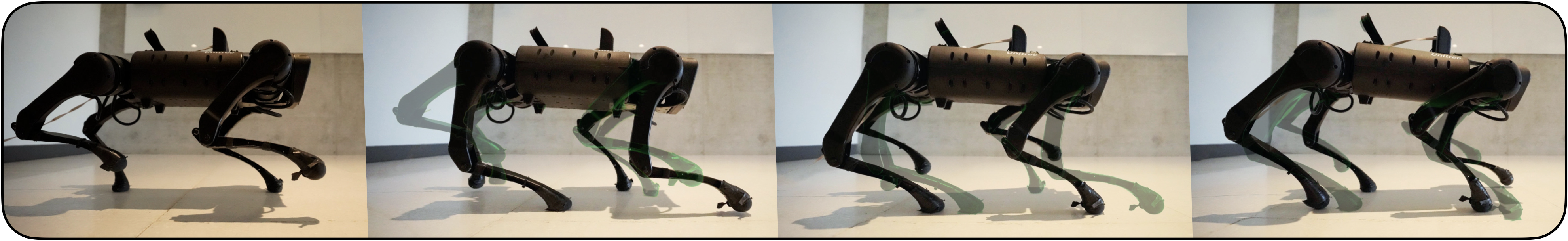}\\
    \vspace{1pt}
    \includegraphics[width=\textwidth]{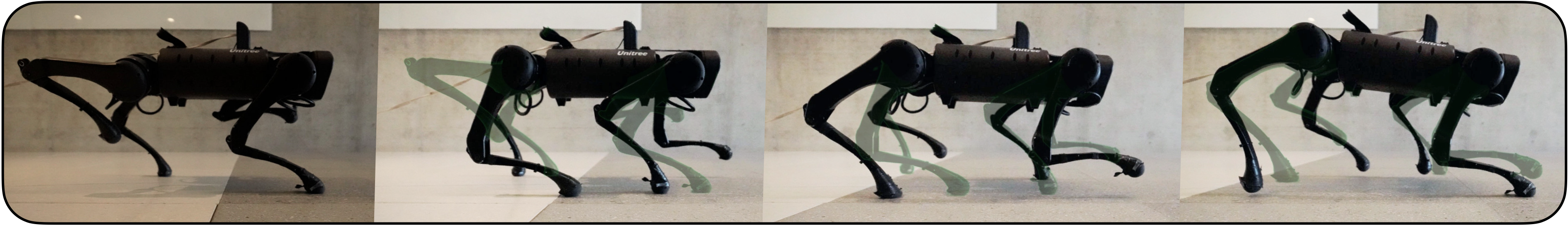}
    \caption{An agent trained with Adversarial Motion Priors extracts the naturalistic locomotion strategies found in the dataset and can change its gait based on the desired velocity. Top: when commanded to move at a low forward velocity $\left(\SI{0.8}{\meter\per\second}\right)$, the agent select a pacing gait. Bottom: when the commanded forward velocity increases to $\left(\SI{1.7}{\meter\per\second}\right)$, the agent switches to a trotting gait. The green low-opacity overlaid images show the previous frame for reference.}
    \label{fig:stills}
\end{figure*}

\section{Method}

\subsection{Background and Problem Formulation}

We model the problem of learning legged locomotion as a Markov Decision Process (MDP): $ (\mathcal{S}, \mathcal{A}, f, r_t, p_0, \gamma)$, where $\mathcal{S}$ is the state space,  $\mathcal{A}$ is the action space, $f(s, a)$ is the system dynamics, $r_t(s, a, s')$ is the reward function, $p_0$ is the initial state distribution, and $\gamma$ is the discount factor. The goal of Reinforcement Learning (RL) is to find the optimal parameters $\theta$ of a policy $\pi_\theta: \mathcal{S} \mapsto \mathcal{A}$ to maximize the expected discounted return $J(\theta) = \mathbb{E}_{\pi_\theta}\left[\sum_{t=0}^{T-1} \gamma^t r_t\right]$.

In this work, we are interested in learning a locomotion controller that is agile and controllable. As such, we design a task reward function for encouraging the robot to track a command velocity $\vec{v}_t = [v_t^x, v_t^y, \omega_t]$ at time $t$, where $v_t^x$ and $v_t^y$ are the desired forward and lateral velocities specified in the base frame, and $\omega_t$ is a desired global yaw rate. In particular, we define the reward function to be:


\begin{equation}
    \label{eq:joystick_control}
    r_t^g = w^v \text{exp}(-{\|\hat{\vec{v}}^\text{ xy}_t - \vec{v}^\text{ xy}_t\|}) + w^\omega \text{exp}(-{|\hat{\omega}_t^z - \omega_t^z|})
\end{equation}

\noindent
where $\hat{\vec{v}}^\text{ xy}_t$ and $\hat{\omega}_t^z$ are the desired linear and angular velocity. The desired base forward velocity $v_t^x$, base lateral velocity $v_t^y$, and global yaw rate $\omega_t$ are sampled randomly from the ranges $\SIrange{-1}{2}{\meter\per\second}$, $\SIrange{-0.3}{0.3}{\meter\per\second}$, and $\SIrange{-1.57}{1.57}{\radian\per\second}$, respectively. Training with this reward grants a high degree of controllability over the robot's movement, and causes the resulting controller to exhibit locomotion behaviors at different speeds. However, as we will show in our experiments, training with only the task reward $r_t^g$ can lead to undesired behaviors such as violent vibrations due to the under-specified nature of the reward. To tackle this problem, we will regularize behaviors of the policy using a data-driven motion prior acquired through adversarial imitation learning.

\subsection{Adversarial Motion Priors as Style Rewards}

In the adversarial imitation learning setting, we consider reward functions that consist of a ``style'' component $r_t^s$ and a task component $r_t^g$, such that

\begin{equation}
    r_t = w^g r_t^g + w^s r_t^s.
    \label{eq:total_reward}
\end{equation}

The style reward $r_t^s$ encourages the agent to produce behaviors that have the same style as the behaviors from a reference dataset. Whereas the task reward is specified by the user, the style reward is learned from a dataset of reference motion clips. Formally, we define a discriminator as a neural network parameterized by $\phi$. The discriminator $D_\phi$ is trained to predict whether a state transition $(s, s')$ is a real sample from the dataset or a fake sample produced by the agent. We borrow the training objective for the discriminator proposed in AMP \cite{peng21}:

\begin{equation}
    \begin{split}
        \argmin_\phi \text{  } & \mathbb{E}_{(s, s') \sim \mathcal{D}}\left[(D_\phi(s, s') - 1)^2\right]\\
        +& \mathbb{E}_{(s, s') \sim \pi_\theta(s, a)}\left[(D_\phi(s, s') + 1)^2\right]\\
        +& {\frac{w^{\text{gp}}}{2}} \mathbb{E}_{(s, s') \sim \mathcal{D}}\left[\|\nabla_\phi D_\phi(s, s')\|^2\right],
    \end{split}
    \label{eq:disc_objective}
\end{equation}

\noindent
where the first two terms encourages the discriminator to distinguish whether a given input state transition is from the reference dataset $\mathcal{D}$ or produced by the agent. The least squares GAN formulation (LSGAN) used in Eq. \ref{eq:disc_objective} has been shown to minimize the Pearson divergence $\chi^2$ divergence between the reference data distribution and the distribution of transitions produced by the agent. The final term in the objective is a gradient penalty, with coefficient $w_\text{gp}$, which penalizes nonzero gradients on samples from the dataset. The gradient penalty mitigates the discriminator's tendency to assign nonzero gradients on the manifold of real data samples, which can cause the generator to overshoot and move off the data manifold. The zero-centered gradient penalty has been shown to reduce oscillations in GAN training, and improve training stability \cite{Mescheder18}. The style reward is then defined by:
\begin{equation}
    r_t^s(s_t, s_{t+1}) = \max [0, 1 - 0.25(D(s, s') - 1)^2],
    \label{eq:amp_reward}
\end{equation}

\noindent
where an additional offset and scaling is applied to bound the reward in the range [0, 1]. The style reward and the task reward are then combined into the composite reward in Eq. \ref{eq:total_reward}. We optimize the parameters of the policy $\pi_\theta$ to maximize the total discounted return of Eq. \ref{eq:total_reward}, and the parameters of the discriminator $D_\phi$ to minimize the objective presented in Eq. \ref{eq:disc_objective}.



The process of training the policy and discriminator is shown in Fig. \ref{fig:algorithm}. First, the policy takes a step in the environment to produce a state transition $(s, s')$. This state transition is fed to the discriminator $D_\phi(s, s')$ to obtain the style reward $r_t^s$. The state transition is also used to compute the task reward $r_t^g$. Finally, the combined reward and states from the environment and reference motion dataset are used to optimize the policy and discriminator.

\subsection{Motion Capture Data Preprocessing}

The raw motion capture data is a time-series of keypoints corresponding to various frames in the subject's motion. In this work, we use the German Shepherd motion capture data provided by \emph{Zhang and Starke et al.} \cite{zhang18}. The dataset consists of short clips of a German Shepherd pacing, trotting, cantering, and turning in place, with a total duration of 4.5 seconds. We follow the process described by \emph{Peng et al.} \cite{peng18} to retarget the German Shepherd motion to the morphology of the A1 quadrupedal robot. We use inverse kinematics to obtain the joint angles and compute the end-effector positions using forward kinematics. We compute the joint velocities, base linear velocities, and angular velocities using finite differences. These quantities define the states in the motion capture dataset $\mathcal{D}$. State transitions are sampled from $\mathcal{D}$ to serve as real samples for training the discriminator. When training in simulation, we also use reference state initialization \cite{peng18} at the start of each episode to initialize the agent from states randomly sampled from $\mathcal{D}$.

\subsection{Model Representation}

We parametrize the policy as a shallow MLP with hidden dimensions of size [512, 256, 128] and exponential linear unit activation layers. The policy outputs both the mean and standard deviation of the output distribution from which target joint angles are sampled. The standard deviation is initialized to $\sigma_i=0.25$.
The policy is queried at 30Hz, and the target joint angles are fed to PD controllers which compute the motor torques.
The policy is conditioned on an observation $o_t$ derived from the state, which contains the robot's joint angles, joint velocities, orientation, and previous actions. The discriminator is an MLP with hidden layers of size [1024, 512] and exponential linear unit activation layers.

\begin{figure}[t]
    \centering
    \includegraphics[width=0.485\textwidth]{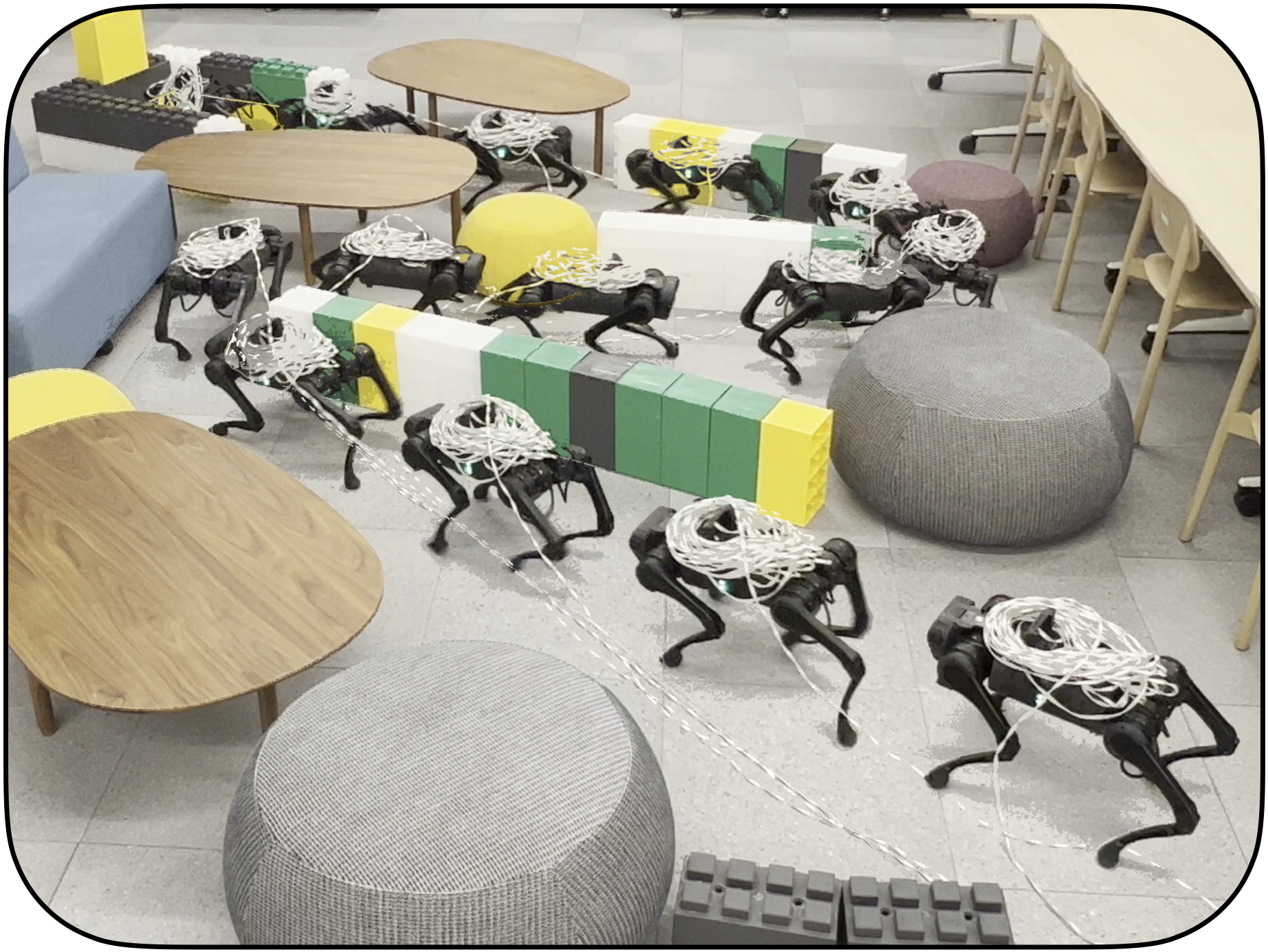}
    \caption{By using Adversarial Motion Priors, the policy can deviate from the reference motion data to satisfy the desired velocity commands and navigate carefully through a route with sharp turns.}.
    \label{fig:overhead}
\end{figure}

\subsection{Domain Randomization} \label{section:domain_rand}

We apply domain randomization to facilitate transfer of learned behaviors from simulation to the real world\cite{tobin17}. Namely, we randomize the terrain friction, base mass, PD controller gains, and perturb the robot's base velocity by adding a sampled velocity vector to the current base velocity at random intervals during training.
The randomization variables and the range of the uniform distribution from which they are sampled are shown in Table \ref{table:domain_rand}.

\begin{table}[h]
\caption{Randomized simulation parameters.}
\label{table:domain_rand}
\begin{center}
\def\arraystretch{1.5}
\begin{tabular}{|c|c|}
\hline
Parameter & Randomization Range\\
\Xhline{4\arrayrulewidth}
Friction & $[0.35, 1.65]$\\
\hline
Added Base Mass & $[-1.0, 1.0]$ kg.\\
\hline
Velocity Perturbation & $[-1.3, 1.3]$ m/s\\
\hline
Motor Gain Multiplier & $[0.85, 1.15]$ \\
\hline
\end{tabular}
\end{center}
\end{table}

\subsection{Training}
We use a distributed PPO\cite{Schulman17} implementation across 5280 simulated environments in Isaac Gym \cite{Liang18, Rudin21}. The policy and discriminator are trained for ~4 billion environment steps, constituting approximately 4.2 years worth of simulation data gathered over 16 hours on a single Tesla V100 GPU. For each training iteration, we collect a batch of 126,720 state transitions $(s, s')$ and optimize the policy and discriminator for 5 epochs with minibatches containing 21,120 transitions. We automatically tune the learning rate to maintain a desired KL divergence of $\text{KL}^\text{desired} = 0.01$ using adaptive LR scheme proposed by \emph{Schulman et al.} \cite{Schulman17}.

The discriminator is optimized with the supervised learning objective in Eq. \ref{eq:disc_objective}. We use the Adam optimizer and a gradient penalty weight of $w^\text{gp}=10$. The style reward and task reward weights are $w^s=0.65$ and $w^g=0.35$, respectively.

\section{Experiments}

In this section we perform quantitative and qualitative analysis of policies trained using various style reward formulations. Namely, we compare the complex reward introduced by \emph{Rudin et al.} \cite{Rudin21} (which is similar to the reward formulations used in other state-of-the-art systems \cite{miki2022learning, lee20, kumar21}) and our approach, which learns a style reward from 4.5 seconds of German Shepherd Motion Capture data. We also include an analysis of policies trained with no style reward. Policies trained with no style reward are analyzed solely in simulation, as the behaviors exhibited by these policies are too violent to deploy on a real robot (Fig. \ref{fig:torque_vel_no_reward}). We seek to answer the following questions:

\begin{itemize}
    \item Do policies trained with Adversarial Motion Priors achieve similar task performance as policies trained with complex style reward formulations?
    \item How energy-efficient are policies trained with the various style reward formulations?
    \item What is the qualitative performance of policies trained with Adversarial Motion Priors when deployed in the real world?
\end{itemize}

\subsection{Task Completion and Energy Efficiency in Simulation} \label{section:sim}

First, we train policies for a velocity tracking task (Eq. \ref{eq:joystick_control}), where the goal is for a policy to closely track the target velocity specified by the user. Here, we compare the performance of policies trained with three reward functions in simulation: no style reward (task reward only), the adversarial motion prior style reward in Eq. \ref{eq:amp_reward}, and the complex style reward proposed by \emph{Rudin et al.} \cite{Rudin21}. The complex style reward is composed of 13 style terms, most of which are designed to penalize behaviors that emerge from an under-specified reward function. Each reward component and the associated scaling factors are listed in the appendix (Table \ref{table:complex_reward}).

We also estimate the Cost of Transport (COT) for each of these policies. COT is a dimensionless quantity commonly used in the field of legged locomotion, as it allows for energy-efficiency comparisons of dissimilar robots or controllers. We utilize the COT to measure the efficiency of different baselines at different speeds. We define the mechanical COT as: $\frac{\textrm{Power}}{\textrm{Weight} \times \textrm{Velocity}} = \sum_{\text{actuators}} [\tau \dot{\theta}]^{+} / (W \|v\|)$, where $\tau$ is the joint torque, $\dot{\theta}$ is the motor velocity, $W$ is the robot's weight, and $\|v\|$ is the velocity.

\begin{figure}[h]
    \centering
    \includegraphics[width=0.48\textwidth]{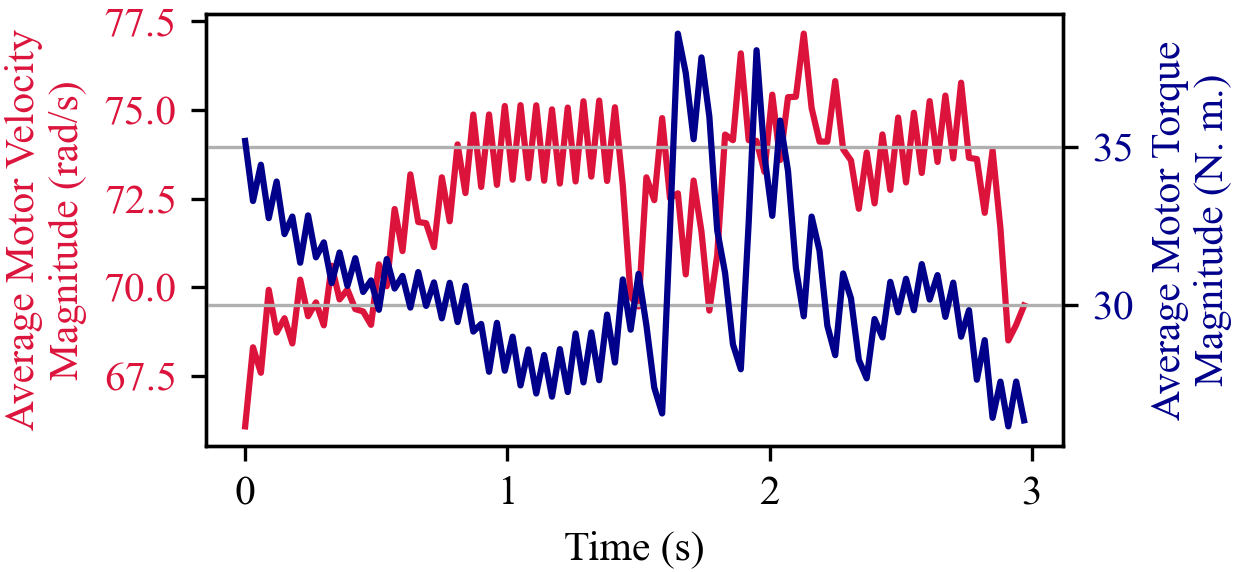}
    \caption{The policy trained with no style reward learns to exploit inaccurate simulator dynamics and violently vibrates the simulated robot's feet on the ground to move. The high motor velocities and torques make it impossible to deploy this control strategy on the real robot.}
    \label{fig:torque_vel_no_reward}
\end{figure}

We find that a policy trained using AMP successfully tracks the desired forward velocity commands while exhibiting a lower COT than competing baselines (Table \ref{table:performance}). Meanwhile, a policy trained with no style reward learns to move by vibrating its legs at high speeds with large torques (Fig. \ref{fig:torque_vel_no_reward}), producing high-impulse contacts with the ground. While this behavior leads to high tracking accuracy, applying such a control strategy to a real robot is infeasible due to the violent nature of the resulting motions and risk of damaging the robot. As shown in Table \ref{table:performance}, policies trained with the hand-designed style reward exhibit a relatively low COT varying between $1.37$ and  $1.65$. Our method produces a lower COT, varying between $0.93$ and  $1.12$ for different target speeds. Meanwhile, a method trained with no style reward exhibits an extremely high COT due to the high torques and motor velocities that emerge from the jittery behaviors.

The energy efficiency of policies trained with AMP can likely be attributed to the policy extracting energy-efficient motion priors from the reference data. Millions of years of evolution has endowed dogs with energy-efficient locomotion behaviors. Training with AMP enables the policy to extract some of these energy-efficient strategies from the data. Additionally, animals often perform gait transitions when undergoing large changes in velocity, lowering the cost of transport across different speeds \cite{Hoyt1981}. The same principle applies to policies trained using AMP. In Fig. \ref{fig:banner_canter}, we see that the robot transitions from a pacing motion to a canter motion when the desired velocity jumps from 1 m/s to 2 m/s. While pacing is the optimal gait at low speeds, entering a canter motion with a flight phase is a more energy-efficient option at high speeds.

\begin{figure}[ht]
    \centering
    \includegraphics[width=0.48\textwidth]{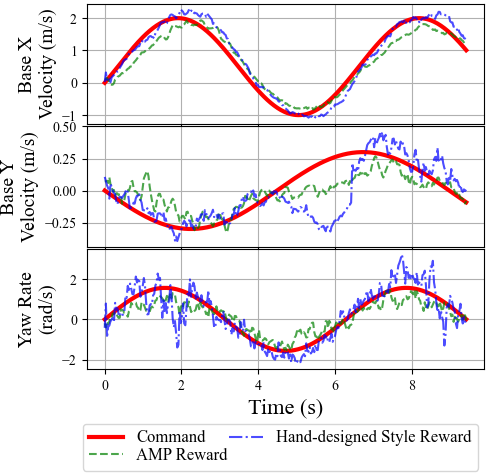}
    \caption{Comparison of motion prior style reward, hand-designed style reward, and no style reward in ability to track a sinusoidal linear and angular velocity command. The policy trained with no style reward was evaluated in simulation due to the violent and jittery behaviors it exhibited (shown in Fig. \ref{fig:torque_vel_no_reward}).}
    \label{fig:tracking}
\end{figure}

\begin{table}[h]
\caption{Velocity tracking and mechanical efficiency.}
\label{table:performance}
\begin{center}
\def\arraystretch{1.6}
\begin{tabular}{|c|c|c|c|c|c|}
\hline
\multicolumn{2}{|c|}{\thead{Commanded Forward \\ Velocity (m/s)}} & 0.4 & 0.8 & 1.2 & 1.6 \\
\Xhline{4\arrayrulewidth}
\multirow{3}{*}{\thead{Average\\Measured\\Velocity\\(m/s)}} & AMP Reward & \thead{$\textbf{0.36}$\\$\pm\textbf{0.01}$} & \thead{$\textbf{0.77}$\\$\pm\textbf{0.01}$} & \thead{$\textbf{1.11}$\\$\pm\textbf{0.01}$} & \thead{$\textbf{1.52}$\\$\pm\textbf{0.03}$} \\
\cline{2-6}
& \thead{Complex\\Style Reward} &\thead{$0.41$\\$\pm0.01$} &\thead{$0.88$\\$\pm0.02$} &\thead{$1.28$\\$\pm0.03$} &\thead{$1.67$\\$\pm0.03$} \\
\cline{2-6}
& \thead{No Style\\Reward} &\thead{$0.42$\\$\pm0.01$} &\thead{$0.82$\\$\pm0.01$} &\thead{$1.22$\\$\pm0.01$} &\thead{$1.61$\\$\pm0.01$} \\
\Xhline{4\arrayrulewidth}
\multirow{3}{*}{\thead{Average\\Mechanical\\Cost of\\Transport}} & AMP Reward &
\thead{$\textbf{1.07}$\\$\pm\textbf{0.05}$} & \thead{$\textbf{0.93}$\\$\pm\textbf{0.04}$} &
\thead{$\textbf{1.02}$\\$\pm\textbf{0.05}$}&
\thead{$\textbf{1.12}$\\$\pm\textbf{0.1}$}\\
\cline{2-6}
& \thead{Complex\\Style Reward} &\thead{$1.54$\\$\pm0.17$} &\thead{$1.37$\\$\pm0.12$} &\thead{$1.40$\\$\pm0.10$} &\thead{$1.41$\\$\pm0.09$} \\
\cline{2-6}
& \thead{No Style\\Reward} &\thead{$14.03$\\$\pm0.99$} &\thead{$8.00$\\$\pm0.44$} &\thead{$6.05$\\$\pm0.28$} &\thead{$5.18$\\$\pm0.20$} \\
\hline
\end{tabular}
\end{center}
\end{table}

\subsection{Task Completion in Real}

Imitation learning approaches that constraint the policy to explicitly track a specified reference motion \cite{peng18} would make it
difficult for the policy to produce behaviors that deviate significantly from the reference data, even though deviating from the reference data may be required to complete the desired task. As shown in Fig. \ref{fig:tracking}, a policy trained using the AMP is able to track the commanded linear and angular velocities, even though the 4.5 second German Shepherd dataset does not contain motions of the dog moving at these particular velocities. This indicates that Adversarial Motion Priors enable the policy to learn behaviors that capture the essence of the reference motions while deviating from the dataset enough to complete the specified task. Take for example Fig. \ref{fig:overhead}: navigating the quadruped through a route with sharp requires precision in tracking the velocity commands. A policy trained with Adversarial Motion Priors can learn to accurately track the velocity commands that guide the robot dog through the course while exhibiting naturalistic locomotion strategies.


\subsection{Qualitative Performance of Policies in Real}

In addition to evaluating the performance of the policies in simulation, we also evaluate the effectiveness of the policies when deployed on a real robot. As mentioned in Section \ref{section:sim}, a compelling benefit of training with Adversarial Motion Priors is that the policy can learn to extract the energy-efficient motion priors from the data. Part of this energy efficiency stems from the policy learning to change its gait depending on the velocity commands. Figure \ref{fig:banner_canter} demonstrates this phenomenon in practice. When the velocity command increases from 1 m/s to 2 m/s, the policy dramatically alters its gait from a pace to a canter. The pacing motion (Fig \ref{fig:stills}:A) is often used by animals at low speed and involves alternating swing and stance phases for the left and right feet. Meanwhile, the canter motion used by animals traveling at high speeds and is composed of an alternating placement of front feet and hind feet, followed by a flight phase (Fig. \ref{fig:banner_canter}:A-2). Transitioning to a canter maneuver results in a dramatically different COT profile (Fig. \ref{fig:banner_canter}:D). While the pace motion exhibits a fairly constant COT, the canter motion produces large spikes in COT corresponding to the lift-off phase and relatively low-valued troughs associated with the flight and touch-down phase. Also shown in Figure \ref{fig:stills}:B is the trotting motion that emerges from training with AMP rewards.


\section{Conclusions}

We demonstrate that learning motion priors using adversarial imitation learning produces style rewards that encourage the policy to produce behaviors that are grounded in the reference motion dataset. Using this technique, we circumvent the need to define complex hand-designed style rewards while still enabling transfer to the real world. Additionally, we demonstrate that policies trained with Adversarial Motion Priors can deviate from the motions in the reference dataset as needed to achieve the specified task objectives. We also compare the energy efficiency of policies trained with hand-defined style reward, AMP style rewards, and no style reward, and demonstrate that AMP style rewards lead to energy-efficient locomotion strategies. We argue that this stems from the energy-efficient prior provided by motions in the dataset, as well as the policy's ability to transition between the most optimal gaits corresponding to the commanded velocities.




\section*{APPENDIX}

\begin{table}[ht]
\caption{Complex reward formulation Baseline.}
\label{table:complex_reward}
\begin{center}
\def\arraystretch{1.5}
\begin{tabular}{|c|c|c|}
    \hline
    Reward Term & Definition & Scale\\
    \Xhline{4\arrayrulewidth}
    z base linear velocity & $(v_t^\text{z})^2$ & -2\\
    \hline
    xy base angular velocity & $\|\vec{\omega}_t^\text{xy}\|$ & -0.05\\
    \hline
    Non-flat base orientation & $\|\vec{g}_t^{\text{ xy}}\|$ & -0.01\\
    \hline
    Torque penalty & $\|\vec{\tau}\|$ & -1e-5\\
    \hline
    DOF acceleration penalty & $\|\ddot{\vec{\theta}}\|$ & -2.5e-7\\
    \hline
    Penalize action changes & $\|\vec{a}_t - \vec{a}_{t-1}\|$ & -0.01\\
    \hline
    Collision penalty & $|I_\text{c, body} \backslash I_\text{c, foot}|$ & -1\\
    \hline
    Termination penalty & $\mathbb{I}_\text{terminate}$ & -0.5\\
    \hline
    DOF lower limits & $-\max(\vec{\theta} - \vec{\theta}_\text{lim, low}, 0)$ & -10.0\\
    \hline
    DOF upper limits & $\min(\vec{\theta} - \vec{\theta}_\text{lim, high}, 0)$ & -10.0\\
    \hline
    Torque limits & $\min(|\vec{\tau} - \vec{\tau}_\text{ lims}|, 0)$ & -0.0002\\
    \hline
    Tracking linear vel & $exp(-{\|\hat{\vec{v}}^\text{ xy}_t - \vec{v}^\text{ xy}_t\|})$ & 1.0\\
    \hline
    Tracking angular vel & $exp(-{|\hat{\omega}_t^z - \omega_t^z|})$ & 0.5\\
    \hline
    Reward long footsteps & $\sum_{\text{feet}} \mathbb{I}_\text{swing} t_\text{swing}$ & 1.0\\
    \hline
    Penalize large contact forces & $\|\min(\vec{f} - \vec{f}_\text{max}, 0)\|$ & -1.0\\
    \hline
\end{tabular}
\end{center}
\end{table}

\section*{ACKNOWLEDGMENT}

The authors would like to thank Adam Lau, Justin Kerr, Lars Berscheid, and Chung Min Kim for their helpful contributions and discussions.


\bibliographystyle{IEEEtran}
\bibliography{root}

\end{document}